\documentclass[11pt]{article}

\usepackage[preprint]{acl}
\usepackage[framemethod=TikZ]{mdframed}
\usepackage{times}
\usepackage{latexsym}
\usepackage{fvextra}
\usepackage{algorithm}
\usepackage{algorithmic}
\usepackage[T1]{fontenc}

\usepackage[utf8]{inputenc}

\usepackage{microtype}

\usepackage{inconsolata}

\usepackage{graphicx}

\usepackage{times}
\usepackage{latexsym}
\usepackage{amsmath,amssymb,amsthm}
\usepackage{booktabs}
\usepackage{multirow}
\usepackage{graphicx}
\usepackage{xcolor}
\usepackage{algorithm}
\usepackage{algorithmic}
\usepackage{microtype}
\usepackage{url}
\usepackage{subcaption}

\newtheorem{definition}{Definition}

\newcommand{\model}{\textsc{HazDial}}

\title{Enhancing Operational Safety via Agentic Dialogue\\Hazard Identification Analysis}

\author{
    \normalsize{Sanjay Das}\\
  \normalsize{Oak Ridge National Laboratory} \\
  \normalsize{Oak Ridge, Tennessee, USA} \\
  \normalsize{\texttt{dass3@ornl.gov}} \\\And
  \normalsize{Ran Elgedawy} \\
  \normalsize Oak Ridge National Laboratory \\
  \normalsize Oak Ridge, Tennessee, USA \\
  \normalsize\texttt{elgedawyr@ornl.gov} \\\And
  \normalsize Ethan Seefried \\
  \normalsize Oak Ridge National Laboratory \\
  \normalsize Oak Ridge, Tennessee, USA \\
  \normalsize \texttt{seefriedej@ornl.gov} \\ \AND %
  \normalsize Ryan Burchfield \\
  \normalsize Oak Ridge National Laboratory \\
  \normalsize Oak Ridge, Tennessee, USA \\
  \normalsize \texttt{burchfieldra@ornl.gov} \\\And 
  \normalsize Tirthankar Ghosal \\
  \normalsize Oak Ridge National Laboratory \\
  \normalsize Oak Ridge, Tennessee, USA \\
  \normalsize \texttt{ghosalt@ornl.gov} }

\begin{document}
    

    
\maketitle
\begin{abstract}
Operational safety in high-stakes domains---such as industrial process control, autonomous, and safety-critical systems---demand reliable hazard identification. While large language models (LLMs) have shown promise in automating safety analysis tasks, single-turn, monolithic inference is brittle: it lacks the self-correction, deliberation, and contextual refinement that safety engineers apply iteratively. In this paper, we introduce \model{}, a framework that investigates whether structured \emph{agentic dialogue}---multi-agent, multi-turn interactions---improves the quality of NLP-based hazard identification over single-pass baselines. We systematically compare two dialogue modalities: \emph{adversarial debate} and \emph{constructive discussion}, and propose an algorithm-based agentic interaction optimization. We evaluate all configurations against a curated golden dataset using standard classification metrics (accuracy, precision, recall, F$_1$) and  novel dialogue metrics. 
This work advances the intersection of dialogue systems, multi-agent reasoning, and AI safety, providing an empirical evidence for dialogue-driven hazard analysis.

\end{abstract}

\section{Introduction}
\label{sec:intro}

Operational safety analysis is a cornerstone of engineering practice. Processes such as Hazard and Operability Studies (HAZOP) \cite{kletz1999hazop}, Failure Mode and Effects Analysis (FMEA) \cite{stamatis2003fmea}, and Fault Tree Analysis (FTA) \cite{vesely1981fta} require safety engineers to systematically enumerate potential hazards, failure modes, their causes, and their consequences. These analyses are labor-intensive, domain-specific, and prone to human errors, fatigue, making them compelling targets for NLP-assisted automation \cite{brown2020gpt3,wei2022chain}.

Recent work has begun to apply LLMs to safety-related text classification and extraction tasks \cite{paltrinieri2019nlp,rajpurkar2018squad}, but a persistent challenge remains: a single-pass LLM inference is epistemically flat. It cannot challenge its own assumptions, incorporate counterfactual reasoning, or refine outputs through deliberation---all of which safety engineers do naturally through structured reviews and peer challenges.

\begin{figure}
    \centering
    \includegraphics[width=\linewidth]{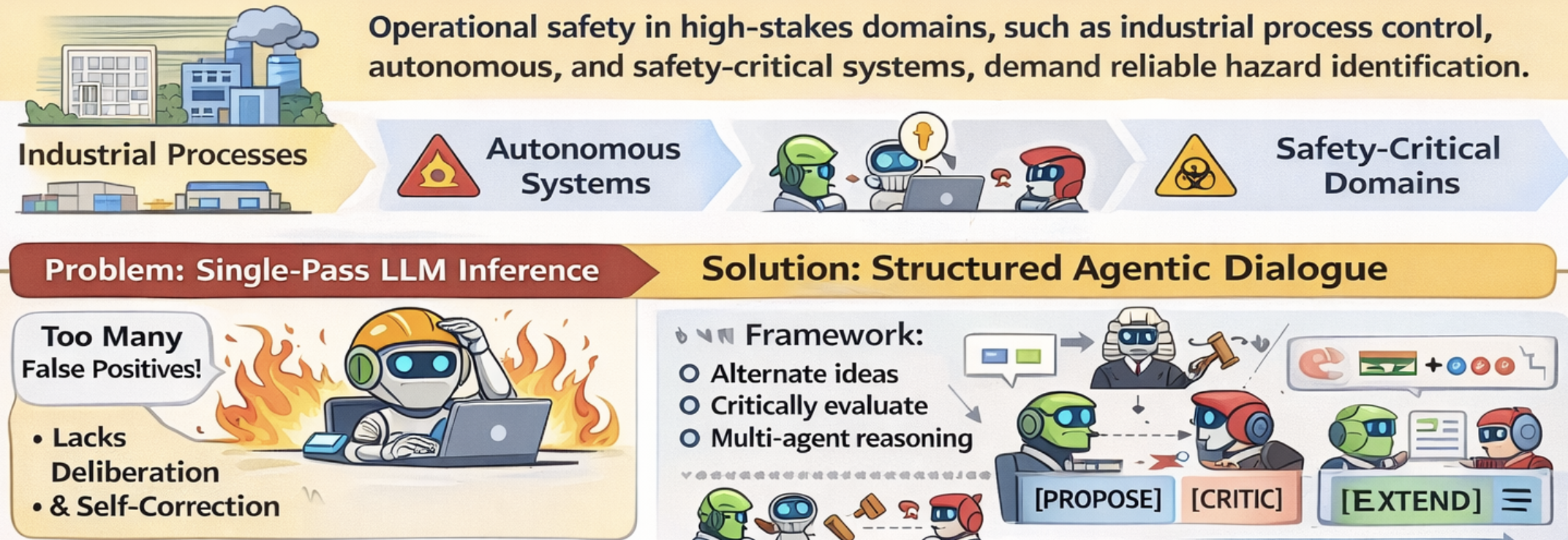}
    \caption{Dialogue-driven hazard analysis.}
    \label{fig:overview}
\end{figure}

\paragraph{Why agentic dialogue?}
Multi-agent dialogue systems introduce structured deliberation into the inference process. When two or more agents interact---whether through cooperative discussion or adversarial debate---the resulting output benefits from error correction, coverage expansion, and iterative refinement \cite{liang2023encouraging,du2023improving}. Yet it remains unclear: (1) whether dialogue \emph{actually} improves hazard identification quality, and (2) what \emph{type} of dialogue structure and \emph{optimization strategy} yields the greatest gains.

\paragraph{The gap we address.}
Prior work on multi-agent debate \cite{du2023improving,chan2023chateval} and collaborative reasoning \cite{wang2023rethinking} has focused primarily on factual QA, mathematical reasoning, and commonsense tasks. Safety-critical hazard identification introduces distinct challenges: outputs must be \emph{comprehensive} (high recall over hazard space), \emph{precise} (low false-positive rate to avoid alarm fatigue), and \emph{explainable} (agents must justify their claims with engineering evidence). Furthermore, the optimization of the \emph{agentic configuration itself}---which agent speaks when, in what order, with what prompt---has received little rigorous attention. We make the following contributions:
\begin{enumerate}
 
  \item We formulate hazard identification as a \emph{closed-list}
        task and define the \emph{Hazard Identification Dialogue} (\model{})
        framework, specifying agents, shared mutable state, dialogue tags,
        and a deterministic aggregation function.
 
  \item We design and implement two structurally distinct multi-agent
        configurations \textsc{Debate} (adversarial Proposer/Critic) and
        \textsc{Discuss} (cooperative Analyst/Reviewer) 
        that prevents circular argumentation and enforces evidence-grounded
        verdicts to study of how dialogue structure
        affects hazard identification.

 \item We propose a genetic policy optimization algorithm that treats the agentic
        configuration parameters as learnable genes, using a recall-prioritized $F_\beta$
        ($\beta = 2$) fitness signal to evolve configurations that
        minimize missed hazards across successive work descriptions.
        \item Experiments with GPT-OSS 20B and GPT-4.1 demonstrate that
        adversarial debate consistently reduces false positives by up
        to 40\% across both models, while constructive discussion
        degrades $F_1$ on the smaller model but achieves the highest
        recall (0.586) on GPT-4.1, showing that cooperative
        dialogue may helps in certain scenarios.
\end{enumerate}

\section{Background and Related Work}
\label{sec:related}

\subsection{Hazard Identification in Safety Engineering}
Hazard identification (HazID) is the first and most critical phase of risk assessment \cite{iec61508,iso26262}. Classical methods---HAZOP, FMEA, STPA \cite{leveson2011stpa}---rely on structured checklists and guidewords (\textit{NO, MORE, LESS, AS WELL AS, REVERSE, OTHER THAN}) applied systematically to process deviations. NLP-based approaches have begun automating aspects of this process: extracting safety-relevant entities \cite{paltrinieri2019nlp}, classifying incident reports, and generating FMEA entries from process descriptions \cite{liao2023fmea}. However, these approaches treat HazID as a static extraction task, ignoring the iterative, collaborative nature of real-world safety reviews.

\subsection{Large Language Models for Safety Analysis}
\citet{brown2020gpt3} and subsequent work have demonstrated broad capability in NLP tasks with minimal supervision. \citet{wei2022chain} introduced chain-of-thought prompting, enabling step-by-step reasoning that is beneficial for complex safety reasoning. \citet{kojima2022large} showed that zero-shot CoT achieves competitive performance, while \citet{yao2023tree} extended this to tree-structured deliberation. Despite these advances, LLMs remain prone to hallucination and overconfidence in safety contexts \cite{ji2023hallucination}, motivating dialogue-based verification.

\subsection{Multi-Agent Dialogue and Debate}
\citet{du2023improving} demonstrated that multi-agent debate---where models critique and revise each other's outputs over multiple rounds---improves factual accuracy and reasoning quality on benchmarks. \citet{liang2023encouraging} showed that diversity of perspectives in multi-agent setups reduces groupthink. \citet{chan2023chateval} proposed ChatEval for reference-free dialogue evaluation using multi-agent frameworks. \citet{xiong2023examining} examined LLM consistency under self-contradiction. In parallel, \citet{park2023generative} explored emergent social behaviors in LLM agent societies. Our work extends this line by (a) applying it to safety-critical HazID, and (b) systematically optimizing the dialogue \emph{configuration itself} through genetic evolutionary optimization.

\subsection{Dialogue Optimization and Policy Learning}
Policy gradient methods for dialogue \cite{li2016deep,williams1992reinforce} and reward shaping \cite{ziegler2019fine} have been applied to task-oriented systems. Our optimization component adapts these ideas to a lightweight, prompt-level policy without full RL, making it computationally tractable for deployment. Evolutionary approaches to prompt optimization \cite{guo2024evoprompt, das2024genbfa} have shown promise; we extend this to the level of entire agent call sequences.

\section{Problem Formulation}
\label{sec:formulation}




\subsection{Hazard Identification}
Let $\mathcal{W}$ denote a corpus of natural-language work descriptions drawn
from operational safety records.  
A hazard identification system $\mathcal{F}$ takes $(w, \mathcal{L})$ as input
and returns a predicted subset $\hat{\mathcal{H}}(w) \subseteq \mathcal{L}$.
 
This framing mirrors real engineering practice: safety analysts consult a
pre-approved hazard taxonomy and decide which entries apply to the work at
hand.  Constraining predictions to $\mathcal{L}$ eliminates unconstrained
generation noise and makes evaluation deterministic, while the selection task
remains non-trivial because $|\mathcal{L}|$ is large and the evidence for each
hazard must be inferred from unstructured prose.
\subsection{Dialogue Modalities}
We study two dialogue modes:

\paragraph{Adversarial Debate (AD).}
Two agents take opposing epistemic stances. Agent $a_1$ (Proposer) identifies hazards; agent $a_2$ (Challenger) critiques and seeks to identify errors, missed hazards, or over-classified events. The Challenger's objective is to \emph{find faults, identify omissions}, and \emph{propose counter-evidence}.

\paragraph{Constructive Discussion (CD).}
Agents collaboratively elaborate hazard coverage from complementary perspectives (e.g., Analyst, Reviewer). Each agent extends, refines, or validates prior contributions without an adversarial objective.
\subsection{Dialogue-Augmented Hazard Identification}
 
\begin{definition}[Hazard Identification Dialogue]
A \emph{Hazard Identification Dialogue} (HID) is a tuple
$\langle \mathcal{A}, T, \Sigma, \Omega \rangle$ where:
\begin{itemize}
  \item $\mathcal{A} = \{a_1, a_2\}$ is a pair of LLM-backed agents, each
        assigned a distinct epistemic role;
  \item $T = (t_1, t_2, \ldots, t_N)$ is an ordered sequence of dialogue
        turns, with $N = 2R$ for $R$ rounds;
  \item $\Sigma$ is a shared mutable \emph{dialogue state} recording which
        hazards have been proposed, accepted, rejected, or withdrawn;
  \item $\Omega : \Sigma \rightarrow \hat{\mathcal{H}}$ is a deterministic
        aggregation function that reads the final state to produce the
        predicted hazard set.
\end{itemize}
\end{definition}
 
At each turn $t_n$, the active agent $a_{i(n)}$ generates a response $r_n$
conditioned on the work description $w$, the master list $\mathcal{L}$, and
the accumulated dialogue history $\mathbf{h}_n = (r_1, \ldots, r_{n-1})$:
\begin{equation}
  r_n = a_{i(n)}\!\left(\pi_{i(n)},\; w,\; \mathcal{L},\; \mathbf{h}_n\right)
  \label{eq:turn}
\end{equation}
where $\pi_{i(n)}$ is the role-specific system prompt for the active agent.
After $R$ rounds, $\Omega$ reads the terminal state $\Sigma_R$ to yield
$\hat{\mathcal{H}}$.
\section{Proposed Systems}
\label{sec:methodology}
 
We investigate three system configurations, progressing from a non-interactive
baseline to increasingly structured multi-agent dialogue.  All three share the
same underlying LLM and the same master hazard list.
 
\subsection{System 1: Single-Prompt Baseline (\textsc{Base})}
\label{sec:base}
 
Before introducing dialogue, we must establish what a competent single-pass
LLM can already achieve.  The baseline mirrors the simplest version: one agent, one turn, no deliberation.
 
\paragraph{Formulation.}
Given work description $w$ and master list $\mathcal{L}$, the agent produces a
predicted hazard set in a single forward pass:
\begin{equation}
  \hat{\mathcal{H}}^{\textsc{Base}}
    = \mathrm{Parse}\!\left(
        a_1\!\left(\pi^{\textsc{base}},\; w,\; \mathcal{L}\right)
      \right)
  \label{eq:base}
\end{equation}
The system prompt $\pi^{\textsc{base}}$ instructs the model to act as a safety
expert and return only hazard labels present verbatim in $\mathcal{L}$.  No
second agent, no state, no iteration.
 
 
\subsection{System 2: Adversarial Debate (\textsc{Debate})}
\label{sec:debate}
 

Prior work has demonstrated that adversarial multi-agent debate improves
factual accuracy on open-domain tasks \cite{du2023improving,liang2023encouraging};
we test whether the same mechanism transfers to the constrained, evidence-bound
domain of hazard identification.
 
\paragraph{Agent roles and dialogue tags.}
\textsc{Debate} employs two agents with opposing epistemic stances:
 
\begin{itemize}
  \item \textbf{Hazard\_Proposer} selects hazards from $\mathcal{L}$ that are
        directly evidenced by $w$ and annotates each with a textual justification.
        Each proposal is tagged \texttt{[PROPOSE]}.
  \item \textbf{Hazard\_Critic} challenges the proposed list, tagging each
        verdict as either \texttt{[AGREE]} (hazard retained) or
        \texttt{[DISAGREE]} (hazard eliminated), with a supporting reason.
\end{itemize}
 
After $R$ rounds of exchange, the Proposer issues a final reconciliation pass,
outputting the agreed-upon set using \texttt{[FINAL]} tags. Full prompt templates are provided in Appendix \ref{app:prompts}.

\paragraph{State system.}
The shared state $\Sigma$ tracks three disjoint sets across rounds:
\begin{equation}
  \Sigma = \bigl(\mathcal{P},\; \mathcal{L}^{+},\; \mathcal{L}^{-}\bigr)
  \label{eq:state_adv}
\end{equation}
where $\mathcal{P}$ is the set of all proposed hazard labels,
$\mathcal{L}^{+}$ (``locked'') contains Critic-agreed hazards, and
$\mathcal{L}^{-}$ (``rejected'') contains Critic-disagreed hazards.
Previously decided hazards are excluded from subsequent Proposer turns to
prevent circular argumentation.
 
 
 
\subsection{System 3: Constructive Discussion (\textsc{Discuss})}
\label{sec:discuss}

Adversarial debate assumes that tension between agents drives quality.
We design \textsc{Discuss} to model a collaborative
safety review pattern,
testing whether cooperative elaboration produces qualitatively different
outcomes to adversarial challenge.
 
\paragraph{Agent roles and dialogue tags.}
\textsc{Discuss} employs two cooperatively-oriented agents:
 
\begin{itemize}
  \item \textbf{Hazard\_Analyst} proposes hazards, tagged \texttt{[SUGGEST]},
        with evidence from $w$.  When queried, the Analyst either confirms a
        hazard (\texttt{[CONFIRM]}) with direct textual evidence or retracts it
        (\texttt{[WITHDRAW]}) when challenged.
  \item \textbf{Hazard\_Reviewer} validates proposals cooperatively: it
        supports well-evidenced hazards (\texttt{[SUPPORT]}), refines imprecise
        labels to more specific entries in $\mathcal{L}$ (\texttt{[REFINE]}),
        or raises a targeted clarification request (\texttt{[QUERY]}) before
        committing.
\end{itemize}
 
This richer tag vocabulary
captures qualitatively different dialogue behaviors: incremental refinement,
epistemic uncertainty, and evidence-contingent commitment.
 
\paragraph{State system.}
The shared state extends Equation~(\ref{eq:state_adv}) with two additional
sets:
\begin{equation}
  \Sigma = \bigl(\mathcal{S},\; \mathcal{C},\; \mathcal{F},\;
                 \mathcal{Q},\; \mathcal{W}\bigr)
  \label{eq:state_dis}
\end{equation}
where $\mathcal{S}$ is the set of suggested hazards, $\mathcal{C}$ the
confirmed set (contributed by \texttt{[SUPPORT]}, \texttt{[REFINE]}, or
\texttt{[CONFIRM]}), $\mathcal{F}$ the set of refinement substitutions
$\{(h_{\mathrm{orig}}, h_{\mathrm{refined}})\}$, $\mathcal{Q}$ the queried
set, and $\mathcal{W}$ the withdrawn set.
 
 
 
\begin{table}[t]
\centering
\caption{Dialogue tag vocabulary by system.  ``Epistemic stance'' indicates
whether the tag conveys agreement, disagreement, or epistemic suspension.}
\label{tab:tags}
\small
\begin{tabular}{llll}
\toprule
\textbf{System} & \textbf{Tag} & \textbf{Agent} & \textbf{Stance} \\
\midrule
\multirow{3}{*}{\textsc{Debate}}
  & \texttt{[PROPOSE]}  & Proposer & Assertion \\
  & \texttt{[AGREE]}    & Critic   & Positive  \\
  & \texttt{[DISAGREE]} & Critic   & Negative  \\
\midrule
\multirow{7}{*}{\textsc{Discuss}}
  & \texttt{[SUGGEST]}  & Analyst  & Assertion  \\
  & \texttt{[SUPPORT]}  & Reviewer & Positive   \\
  & \texttt{[REFINE]}   & Reviewer & Positive+  \\
  & \texttt{[QUERY]}    & Reviewer & Suspended  \\
  & \texttt{[CONFIRM]}  & Analyst  & Positive   \\
  & \texttt{[WITHDRAW]} & Analyst  & Negative   \\
  & \texttt{[PASS]}     & Either   & Neutral    \\
\bottomrule
\end{tabular}
\end{table}

\subsection{System 4: Evolutionary Policy Optimization (\textsc{GA-Debate})}
\label{sec:ga}
 
Both previous dialogue systems expose a discrete configuration space:
how many rounds to run, how aggressively the Proposer should generate,
how strictly the Critic should filter, and at what level of
chain-of-thought verbosity.  Manually tuning these parameters is
impractical at scale.  We propose a \emph{Genetic Algorithm} (GA) that
learns the optimal configuration by treating each parameter assignment
as an individual and using prediction performance as fitness.
 
\paragraph{Learnable parameters.}
Five parameters constitute an individual $\mathbf{\theta}$:
 
\begin{table}[h]
\centering
\small
\resizebox{\columnwidth}{!}{%
\begin{tabular}{lll}
\toprule
\textbf{Param} & \textbf{Description} & \textbf{Values} \\
\midrule
$\theta_1$ & Number of rounds & $\{1,2,3\}$ \\
$\theta_2$ & Proposer persona & Thorough, Focused, Cautious \\
$\theta_3$ & Critic persona & Strict, Balanced, Skeptical \\
$\theta_4$ & Reasoning depth & high, medium, low \\
$\theta_5$ & Proposal cap/round & $\{5, 10, 20\}$ \\
\bottomrule
\end{tabular}}
\caption{GA parameter space $\Theta$.}
\label{tab:params}
\end{table}
 
\paragraph{Fitness function.}
Given the asymmetric cost of missed hazards in safety-critical
contexts, where a false negative (missed hazard) can lead to
injury or incident while a false alarm (false positive) merely
requires investigation, we use the $F_\beta$ score with $\beta = 2$
as the fitness signal~\cite{lee2021surrogate}:
\begin{equation}
  \text{Fit}(\hat{\mathcal{H}},\, \mathcal{H}^{*})
  = \frac{(1+\beta^2)\cdot P \cdot R}{\beta^2 \cdot P + R},
  \quad \beta = 2
  \label{eq:fitness}
\end{equation}
This weights recall four times more heavily than precision in the
harmonic mean, incentivising the GA to converge on configurations that
minimize false negatives even at some precision cost.

\paragraph{GA lifecycle.}
Algorithm~\ref{alg:ga} formalizes the full optimization procedure.
The algorithm operates in two phases and treats each work description
as a single fitness evaluation, enabling continual online learning.
 
\begin{algorithm}[t]
\caption{Evolutionary Policy Optimization for Agentic Hazard Identification}
\label{alg:ga}
\begin{algorithmic}[1]
 
\REQUIRE
  $\mathcal{W} = \{w_1, \ldots, w_N\}$, $\{\mathcal{H}^*(w_i)\}$, $\Theta$, $N_{\mathrm{init}}, K_{\mathrm{elite}}, K_{\mathrm{offs}}$,
  mutation rate $p_{\mathrm{mu}}$
\ENSURE
  Best individual $\boldsymbol{\theta}^*$ and fitness history
  $\mathcal{A}$
\STATE $\mathcal{E} \leftarrow \emptyset$
  \hfill \COMMENT{Evaluated individuals archive}
\STATE $\mathcal{A} \leftarrow \emptyset$
  \hfill \COMMENT{Fitness trajectory log (all sessions)}
\STATE $g \leftarrow 0$
  \hfill \COMMENT{Generation counter}
\FOR{$i = 1$ \TO $N_{\mathrm{init}}$}
  \STATE $\boldsymbol{\theta}_i \leftarrow \textsc{RandomSample}(\Theta)$
  \STATE $\hat{\mathcal{H}}_i \leftarrow
    \textsc{RunDebate}(w_i,\, \mathcal{L},\, \boldsymbol{\theta}_i)$
  \STATE $f_i \leftarrow
    \textsc{Fitness}(\hat{\mathcal{H}}_i,\, \mathcal{H}^*(w_i))$
  \STATE $\mathcal{E} \leftarrow \mathcal{E} \cup
    \{(\boldsymbol{\theta}_i,\, f_i)\}$,\quad
    $\mathcal{A} \leftarrow \mathcal{A} \cup
    \{(\boldsymbol{\theta}_i,\, f_i,\, g)\}$
\ENDFOR
\STATE $\mathcal{P} \leftarrow
  \textsc{BuildPool}(\mathcal{E},\, K_{\mathrm{elite}},\,
    K_{\mathrm{offs}},\, p_{\mathrm{mu}})$
\STATE $j \leftarrow 0$
  \hfill \COMMENT{Pool pointer}
\FOR{$i = N_{\mathrm{init}}+1$ \TO $N$}
  \STATE $\boldsymbol{\theta}^{\mathrm{cur}}
    \leftarrow \mathcal{P}[j \bmod |\mathcal{P}|]$
  \STATE $\hat{\mathcal{H}}_i \leftarrow
    \textsc{RunDebate}(w_i,\, \mathcal{L},\,
    \boldsymbol{\theta}^{\mathrm{cur}})$
  \STATE $f_i \leftarrow
    \textsc{Fitness}(\hat{\mathcal{H}}_i,\, \mathcal{H}^*(w_i))$
  \STATE Update mean fitness of $\boldsymbol{\theta}^{\mathrm{cur}}$
    in $\mathcal{E}$;\quad
    $\mathcal{A} \leftarrow \mathcal{A} \cup
    \{(\boldsymbol{\theta}^{\mathrm{cur}},\, f_i,\, g)\}$
  \STATE $j \leftarrow j + 1$
  \IF{$j \bmod |\mathcal{P}| = 0$}
    \STATE $\mathcal{P} \leftarrow
      \textsc{BuildPool}(\mathcal{E},\, K_{\mathrm{elite}},\,
        K_{\mathrm{offs}},\, p_{\mathrm{mu}})$
    \STATE $g \leftarrow g + 1$;\quad $j \leftarrow 0$
  \ENDIF
\ENDFOR
\STATE $\boldsymbol{\theta}^* \leftarrow
  \arg\max_{\boldsymbol{\theta} \in \mathcal{E}}\;
  \overline{f}(\boldsymbol{\theta})$
\RETURN $\boldsymbol{\theta}^*$, $\mathcal{A}$
\end{algorithmic}
\end{algorithm}

\noindent\textit{Phase 1 — Exploration.}
The first $N_{\mathrm{init}}$ data points are each evaluated with
a distinct randomly sampled individual, covering the parameter space
before any selection occurs (\textit{lines 1-11}).

\noindent\textit{Phase 2 — Evolution.}
After each pool of $K_{\mathrm{elite}} + K_{\mathrm{offs}}$
evaluated individuals, they are ranked by mean fitness. Then top $K_{\mathrm{elite}}$ as the elite set are retained. Next $K_{\mathrm{offs}}$ offspring are generated by applying
gene-wise mutation (rate $p_{\mathrm{mu}}$) to randomly selected elite parents to form the new active pool as elite $\cup$ offspring. This process is continued until all data points are exhausted.

\section{Experimental Setup}
\label{sec:experiments}
 
\subsection{Dataset}
 
We evaluate on the \textbf{HazID-Ops} dataset comprising 213
real-world operational work descriptions drawn from safety management
systems across construction, maintenance, chemical processing, and
electrical infrastructure domains.  Each description is paired with
expert-annotated golden hazard labels drawn from a master list of
standardised entries compiled from domain safety taxonomies.
 
\subsection{Models and Inference Platform}
 
All experiments were conducted on NVIDIA A100 GPU servers.
Two language model configurations are evaluated:
 
\begin{itemize}
  \item \textbf{GPT-OSS 20B} (\texttt{gpt-oss:20b}): an open-source
    20B-parameter instruction-tuned model served locally via Ollama
    with temperature 0.2.  Used for the primary OSS comparison~\cite{agarwal2025gpt}.
  \item \textbf{GPT-4.1} (\texttt{gpt-4.1}): accessed via the OpenAI
    API with temperature 0.2~\cite{wang2025does}.  Used for the GPT comparison and
    GA-optimized experiments.
\end{itemize}
 
Using the same model for all agent roles within a configuration ensures
that performance differences are attributable to dialogue structure and
policy parameters, not model capacity.
 
\subsection{Evaluation Metrics}
\label{sec:metrics}
 
\paragraph{Classification metrics.}
In label matching we use fuzzy Jaccard similarity~\cite{li2021auto} ($\geq 0.60$ threshold)
over morphologically normalized word tokens, tolerating trivial surface
variants (plurals, verb inflections, hyphenation differences).  We
report precision, recall, F1 score and accuracy~\cite{goutte2005probabilistic}.
Macro-averages across scenarios are reported with standard deviations.
 
\paragraph{Dialogue metrics.}
Derived from tagged turn records (Section \ref{sec:methodology}):
\begin{itemize}
  \item $\rho_\mathrm{ag}$: \textbf{Agreement Rate} ---
    fraction of proposals receiving a positive verdict
    (\texttt{[AGREE]}, \texttt{[SUPPORT]}, \texttt{[CONFIRM]}, or
    \texttt{[REFINE]}).
  \item $\rho_\mathrm{dis}$: \textbf{Disagreement/Withdrawal Rate} ---
    fraction of proposals rejected or withdrawn.
  \item $\rho_\mathrm{qry}$: \textbf{Query Rate} ---
    fraction of suggestions queried before commitment
    (\textsc{Discuss} only).
  \item $\rho_\mathrm{ref}$: \textbf{Refine Rate} ---
    fraction of suggestions replaced by a more precise label
    (\textsc{Discuss} only).
  \item $\rho_\mathrm{ss}$: \textbf{Stance Shift Rate} ---
    fraction of rounds in which the confirmed/locked set changed.
  \item $\eta$: \textbf{Dialogue Efficiency} ---
    ratio of final confirmed hazards to total proposals.
  \item \textbf{Convergence Rate}: fraction of scenarios where the
    confirmed set stabilized before the final round.
\end{itemize}
These metrics are defined in Appendix \ref{app:dialogue_metrics}.
\section{Results}
\label{sec:results}
 
\subsection{OSS 20B Model: Base, Debate, and Discuss}
 
Table~\ref{tab:oss_clf} reports classification metrics for all three
systems on the GPT-OSS 20B model across 213 scenarios. Table~\ref{tab:oss_dlg} presents the corresponding dialogue metrics.
 \begin{figure}[t]
  \centering
  \includegraphics[width=\columnwidth]{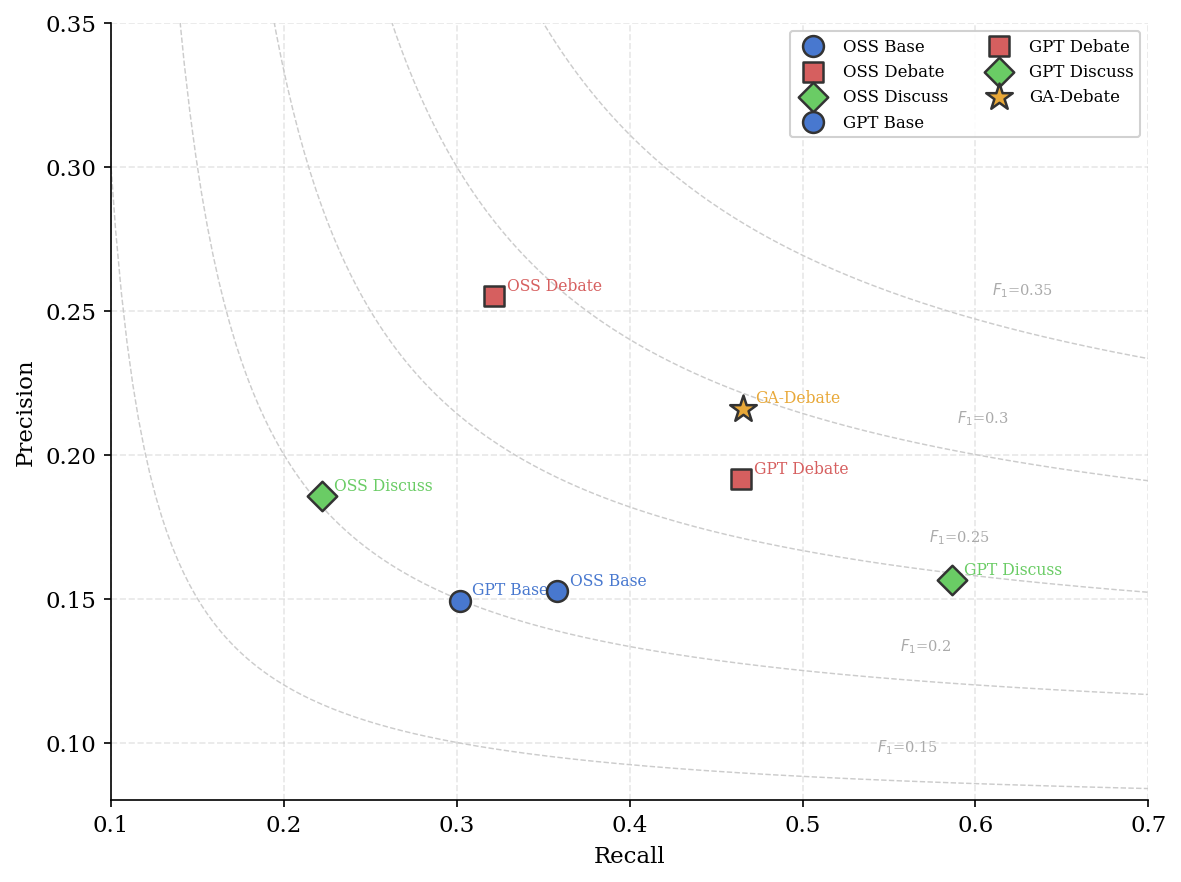}
  \caption{Precision vs.\ recall for all systems.}
  \label{fig:pr_scatter}
\end{figure}
\begin{table}[b]
\centering
\caption{Classification metrics — GPT-OSS 20B.}
\label{tab:oss_clf}
\resizebox{\columnwidth}{!}{%
\begin{tabular}{lcccc}
\toprule
\textbf{System} & \textbf{Precision} & \textbf{Recall}
               & $\mathbf{F_1}$ & \textbf{Accuracy} \\
\midrule
\textsc{Base}
  & 0.1526 & 0.3582 & 0.2140 & 0.1198 \\
\textsc{Debate}
  & 0.2550$\pm$0.2454 & 0.3216$\pm$0.2733
  & \textbf{0.2492}$\pm$0.1992 & \textbf{0.1589}$\pm$0.1485 \\
\textsc{Discuss}
  & 0.1856$\pm$0.2299 & 0.2218$\pm$0.2524
  & 0.1763$\pm$0.1841 & 0.1089$\pm$0.1216 \\
\bottomrule
\end{tabular}%
}
\end{table} 

\begin{figure}[t]
  \centering
  \includegraphics[width=\columnwidth]{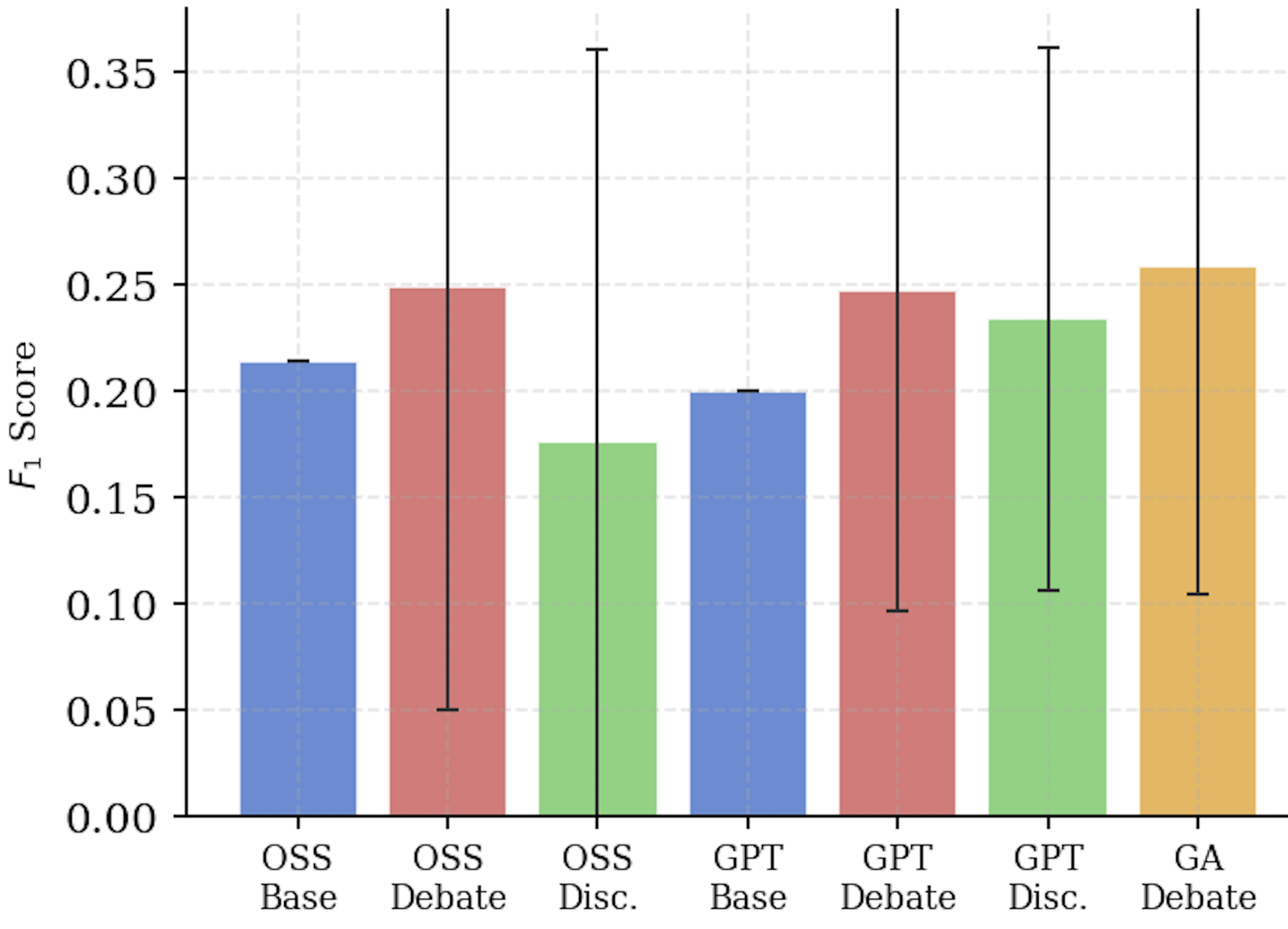}
  \caption{F$_1$ score for all systems.}
  \label{fig:f1}
\end{figure} 

\begin{table}[t]
\centering
\caption{Dialogue metrics — GPT-OSS 20B.}
\label{tab:oss_dlg}
\resizebox{\columnwidth}{!}{%
\begin{tabular}{lcc}
\toprule
\textbf{Metric} & \textbf{\textsc{Debate}} & \textbf{\textsc{Discuss}} \\
\midrule
Agreement Rate $\rho_\mathrm{ag}$
  & 0.5988$\pm$0.2267 & 0.6055$\pm$0.2471 \\
Disagreement/Withdrawal Rate $\rho_\mathrm{dis}$
  & 0.4012$\pm$0.2267 & 0.8998$\pm$0.2572 \\
Query Rate $\rho_\mathrm{qry}$
  & --- & 0.4323$\pm$0.2487 \\
Refine Rate $\rho_\mathrm{ref}$
  & --- & 0.2020$\pm$0.1810 \\
Stance Shift Rate $\rho_\mathrm{ss}$
  & 0.7715$\pm$0.2682 & 0.7121$\pm$0.2899 \\
Dialogue Efficiency $\eta$
  & 0.5918$\pm$0.2269 & 0.5587$\pm$0.2368 \\
Convergence Rate
  & 0.4883 & 0.5915 \\
Mean Convergence Round
  & 2.37$\pm$0.57 & 2.23$\pm$0.67 \\
\bottomrule
\end{tabular}%
}
\end{table}
 
\paragraph{Finding 1: Adversarial debate is the best-performing
OSS configuration.}
\textsc{Debate} improves $F_1$ by $+0.035$ over \textsc{Base}
($0.2492$ vs.\ $0.2140$) and reduces the corpus-level FP count by
40.0\% (from 1{,}761 to 1{,}057).  The improvement is driven entirely by precision
($+0.102$), with a modest recall loss ($-0.037$) attributable to the
Critic's evidence-grounded rejection defaults.
 
\paragraph{Finding 2: Constructive discussion underperforms
the base model.}
Despite a 43.8\% FP reduction, \textsc{Discuss} posts $F_1 = 0.1763$,
below both \textsc{Base} and \textsc{Debate}.  The withdrawal rate of
$\rho_\mathrm{dis} = 0.8998$, meaning that 90\% of queried hazards
are subsequently retracted by the Analyst. The query-withdraw mechanism eliminates true positives alongside false
positives, producing a net recall loss of $-0.137$ relative to
\textsc{Base}.
 
\subsection{GPT-4.1 Model: Base, Debate, and Discuss}
 
Table~\ref{tab:gpt_clf} reports classification metrics for GPT-4.1. Table~\ref{tab:gpt_dlg} presents the  dialogue metrics.

\begin{table}[b]
\centering
\caption{Classification metrics — GPT-4.1.}
\label{tab:gpt_clf}
\resizebox{\columnwidth}{!}{%
\begin{tabular}{lcccc}
\toprule
\textbf{System} & \textbf{Precision} & \textbf{Recall}
               & $\mathbf{F_1}$ & \textbf{Accuracy} \\
\midrule
\textsc{Base}
  & 0.1492 & 0.3018 & 0.1997 & --- \\
\textsc{Debate}
  & 0.1914$\pm$0.1626 & \textbf{0.4644}$\pm$0.2986
  & 0.2473$\pm$0.1513 & 0.1498$\pm$0.1016 \\
\textsc{Discuss}
  & 0.1564$\pm$0.1026 & \textbf{0.5864}$\pm$0.2853
  & 0.2336$\pm$0.1276 & 0.1383$\pm$0.0845 \\
\bottomrule
\end{tabular}%
}
\end{table}

\begin{table}[t]
\centering
\caption{Dialogue metrics — GPT-4.1.}
\label{tab:gpt_dlg}
\resizebox{\columnwidth}{!}{%
\begin{tabular}{lcc}
\toprule
\textbf{Metric} & \textbf{\textsc{Debate}} & \textbf{\textsc{Discuss}} \\
\midrule
Agreement Rate $\rho_\mathrm{ag}$
  & 0.5710$\pm$0.1707 & 1.021$\pm$0.0843$^\dagger$ \\
Disagreement/Withdrawal Rate $\rho_\mathrm{dis}$
  & 0.4289$\pm$0.1703 & 0.5708$\pm$0.4634 \\
Query Rate $\rho_\mathrm{qry}$
  & --- & 0.0729$\pm$0.0607 \\
Refine Rate $\rho_\mathrm{ref}$
  & --- & 0.0967$\pm$0.0556 \\
Stance Shift Rate $\rho_\mathrm{ss}$
  & 0.9208$\pm$0.1566 & \textbf{1.000}$\pm$0.000 \\
Dialogue Efficiency $\eta$
  & 0.5651$\pm$0.1676 & 1.045$\pm$0.0887$^\dagger$ \\
Convergence Rate
  & 0.2178 & 0.000 \\
Mean Convergence Round
  & 2.45$\pm$0.50 & --- \\
\bottomrule
\multicolumn{3}{l}{\small $^\dagger$ Values $>$1 indicate a tag-parsing
  artefact; see \S\ref{sec:discussion}.} \\
\end{tabular}%
}
\end{table}
 
\paragraph{Finding 3: Both dialogue modes substantially improve
recall on GPT-4.1.}
\textsc{Debate} raises recall from $0.3018$ to $0.4644$
($\Delta = +0.163$),
achieving $F_1 = 0.2473$.  \textsc{Discuss} achieves the highest
recall observed across all experiments ($0.5864$)($\Delta = -0.048$), though at a
FP:TP ratio of 5.31, which is only marginally better than the base model's
5.70.
 
\paragraph{Finding 4: GPT-4.1 reverses the OSS behaviour of
\textsc{Discuss}.}
On GPT-OSS~20B, \textsc{Discuss} underperformed \textsc{Base} due to
over-pruning ($\rho_\mathrm{dis} = 0.90$).  On GPT-4.1, the same
architecture achieves the highest recall of any system, with a far
lower withdrawal rate ($0.5708$) and near-zero query rate ($0.073$).
The GPT-4.1 Analyst is more willing to confirm queried hazards,
resulting in larger confirmed sets.  The stance shift rate of $1.000$
(every round changed the confirmed set) and convergence rate of $0.000$
(no scenario converged within 3 rounds) indicate that GPT-4.1
Discuss operates in a regime of continued expansion rather than
selective refinement.
 
\paragraph{Finding 5: Dialogue efficiency anomalies in GPT-4.1
Discuss.}
The acceptance rate of $1.021$ and dialogue efficiency of $1.045$
exceed 1.0, which is mathematically infeasible under the tag
definitions.  This arises from tag-parsing artifacts: GPT-4.1
occasionally outputs multiple \texttt{[SUPPORT]} or \texttt{[SUGGEST]}
tags referencing the same hazard within a single utterance, inflating
the denominator count. 
Classification metrics are computed from parsed
final hazard lists and are therefore unaffected.
 
\subsection{GA-Optimized Adversarial Debate (\textsc{GA-Debate})}
 
Following the GPT-OSS 20B and GPT-4.1 results analysis, \textsc{Debate} was selected for
evolutionary optimization as the best-performing dialogue modality by
$F_1$.  Table~\ref{tab:ga_clf} presents the
classification results, Table~\ref{tab:ga_dlg} the dialogue metrics,
and Table~\ref{tab:ga_policy} the converged optimal policy.
 
\begin{table}[t]
\centering
\caption{Classification metrics — \textsc{GA-Debate} vs GPT-4.1 baselines.}
\label{tab:ga_clf}
\resizebox{\columnwidth}{!}{%
\begin{tabular}{lcccc}
\toprule
\textbf{System} & \textbf{Precision} & \textbf{Recall}
               & $\mathbf{F_1}$ & \textbf{Accuracy} \\
\midrule
GPT-4.1 \textsc{Base}
  & 0.1492 & 0.3018 & 0.1997 & --- \\
GPT-4.1 \textsc{Debate}
  & 0.1914$\pm$0.1626 & 0.4644$\pm$0.2986
  & 0.2473$\pm$0.1513 & 0.1498$\pm$0.1016 \\
\textsc{GA-Debate}
  & \textbf{0.2159}$\pm$0.1901 & \textbf{0.4656}$\pm$0.2502
  & \textbf{0.2585}$\pm$0.1541 & \textbf{0.1579}$\pm$0.1078 \\
\bottomrule
\end{tabular}%
}
\end{table}
 
\begin{table}[b!]
\centering
\caption{Dialogue metrics — \textsc{GA-Debate} vs GPT-4.1 \textsc{Debate}.}
\label{tab:ga_dlg}
\resizebox{\linewidth}{!}{%
\begin{tabular}{lcc}
\toprule
\textbf{Metric} & \textbf{GPT-4.1 Debate} & \textbf{GA-Debate} \\
\midrule
$\rho_\mathrm{ag}$ & 0.5710 & \textbf{0.7939}$\pm$0.2122 \\
$\rho_\mathrm{dis}$ & 0.4289 & \textbf{0.2046}$\pm$0.2114 \\
$\rho_\mathrm{ss}$ & 0.9208 & \textbf{0.9968}$\pm$0.0327 \\
$\eta$             & 0.5651 & \textbf{0.7874}$\pm$0.2411 \\
Convergence Rate   & 0.2178 & 0.0097 \\
Mean Conv.\ Round  & 2.45   & 3.00 \\
\bottomrule
\end{tabular}}
\end{table}
 
\begin{table}[t]
\centering
\caption{Optimal policy discovered by the GA after convergence.}
\label{tab:ga_policy}
\begin{tabular}{ll}
\toprule
\textbf{Parameter} & \textbf{Converged Value} \\
\midrule
$\theta_1$: num\_rounds      & 3 \\
$\theta_2$: proposer\_persona & Thorough \\
$\theta_3$: critic\_persona   & Strict \\
$\theta_4$: reasoning\_depth  & medium \\
$\theta_5$: proposal\_cap     & 20 \\
\bottomrule
\end{tabular}
\end{table}
 
\paragraph{Finding 6: GA optimization improves precision and $F_1$
without sacrificing recall.}
\textsc{GA-Debate} achieves $F_1 = 0.2585$ compared to $0.2473$ for
vanilla GPT-4.1 \textsc{Debate} ($\Delta F_1 = +0.011$) and $0.1997$
for GPT-4.1 \textsc{Base} ($\Delta F_1 = +0.059$).  Recall is
maintained at $0.4656$ (essentially identical to \textsc{Debate}'s
$0.4644$) while precision rises from $0.1914$ to $0.2159$
($\Delta P = +0.025$), demonstrating that the GA improves the
precision--recall balance rather than trading one for the other.
 
\paragraph{Finding 7: GA-optimized dialogue is substantially more
deliberative.}
The agreement rate rises from $0.571$ to $0.794$
($\Delta \rho_\mathrm{ag} = +0.223$), and dialogue efficiency improves
from $0.565$ to $0.787$ ($\Delta \eta = +0.222$), indicating that
the Proposer's suggestions are better calibrated to the Critic's
standards.  The near-unity stance shift rate ($0.997$) confirms active
revision of the hazard set in virtually every round.
 
\paragraph{Finding 8: The converged policy recovers a recall-oriented
configuration.}
The \texttt{Thorough} Proposer persona (which prioritizes coverage over
precision) combined with a high proposal cap of 20 and a
\texttt{Strict} Critic reflects the fitness function's recall bias
($\beta = 2$): the GA learned to generate broadly and filter
aggressively, rather than generating conservatively.  The
\texttt{medium} reasoning depth balances deliberation quality against
inference cost.
 
\section{Discussion}
\label{sec:discussion}
 
\subsection{The Asymmetric Value of Dialogue Across Model Capacity}
 
The most striking cross-model finding is that dialogue mode interacts
strongly with model capacity in determining the direction of
improvement.  On GPT-OSS 20B, adversarial debate improves $F_1$
while constructive discussion harms it; on GPT-4.1, both modes
improve $F_1$ and constructive discussion achieves the highest recall
observed.  We attribute this to the quality of evidence grounding: the
\texttt{[QUERY]}/\texttt{[CONFIRM]}/\texttt{[WITHDRAW]} protocol in
\textsc{Discuss} requires the Analyst to defend suggestions with
specific textual evidence.  GPT-OSS 20B frequently withdraws valid
hazards when challenged because it cannot generate sufficiently
specific justifications; GPT-4.1's stronger language grounding allows
it to confirm borderline hazards correctly, flipping the withdrawal
rate from $0.90$ to $0.57$.  This suggests that cooperative discussion
 architectures may be weak in smaller models. 
 
 
 
\subsection{Evolutionary Policy Learning Aligns the Proposer--Critic
Calibration}
 
The GA's most substantive finding is the emergence of the
\texttt{Thorough + Strict} persona combination.  Intuitively, a
Thorough Proposer with an higher-cap ($= 20$) generates more
candidates, ensuring recall is not sacrificed in the proposal phase;
the Strict Critic then filters aggressively, recovering precision.
This pipeline structure mirrors a ``generate-then-verify'' paradigm
that the system discovered without being explicitly programmed.  The
agreement rate improvement ($+0.223$) indicates that after
optimization, Proposer outputs are substantially better matched to the
Critic's evidence standards, reducing wasted computation on proposals
that will be rejected.
 
\subsection{Dialogue Efficiency as a Convergence Indicator}
 
The GA-optimized system's convergence rate drops to $0.010$ (from
$0.218$), meaning almost no scenario converges before round 3.  This
initially appears negative but reflects the Thorough Proposer
continually surfacing new hazards each round under the high cap,
maintaining productive coverage expansion throughout.  Combined with
the high stance shift rate ($0.997$), this indicates that all three
rounds contribute meaningfully to hazard discovery, validating the
 GA's selection of $\theta_1 = 3$ rounds.
 
\subsection{Limitations}
 
 
\paragraph{Tag-parsing artefacts.}
GPT-4.1 \textsc{Discuss} produces acceptance rates and dialogue
efficiency scores above 1.0 due to repeated tag emission within single
utterances.  Robust tag deduplication should be applied in future work
to make these metrics directly comparable across models.
 
\paragraph{GA evaluation cost.}
The exploration phase requires $N_{\mathrm{init}} = 20$ distinct policy
evaluations before selection begins.  For larger datasets, the
exploration overhead is proportionally smaller; for small datasets, a
warm-start from prior domain knowledge could reduce it.
 
\paragraph{Fuzzy matching threshold.}
The Jaccard threshold of 0.60 handles most surface variants but does
not capture derivational morphology (e.g.\ \textit{electric} vs.\
\textit{electrical}).  An embedding-based matching layer would improve
evaluation fidelity for label vocabularies with high morphological
diversity.
 
\section{Conclusion}
\label{sec:conclusion}
 
This paper presented a systematic investigation of agentic dialogue for
operational hazard identification from a closed label list.  We
evaluated three base configurations---single-prompt inference
(\textsc{Base}), adversarial debate (\textsc{Debate}), and cooperative
discussion (\textsc{Discuss})---across two language models (GPT-OSS 20B
and GPT-4.1), and proposed an evolutionary policy optimization
framework (\textsc{GA-Debate}) that learns the optimal agentic
configuration from prediction feedback. The results demonstrates the concrete gains of an agentic dialogue system on hazard identification performance.

\section{Acknowledgment}
This research is sponsored by the Office of the Laboratory Director, Oak Ridge National Laboratory's Operational Excellence Initiatives, which is supported by the United
States Department of Energy (DOE)’s Office of Science under Contract No. DE-AC05-00OR22725. 
\bibliography{references}
\newpage
\appendix

\appendix

\section{Dialogue Metrics: Definitions and Interpretation}
\label{app:dialogue_metrics}

This appendix provides formal definitions of all dialogue metrics
reported in Section \ref{sec:metrics}, together with interpretive notes on
boundary conditions and known measurement artifacts.




\subsection{Formal Metric Definitions}

Let $R$ denote the number of dialogue rounds per session,
$N_{\textsc{prop}}$ the total number of proposals or suggestions
across all rounds, and subscripts $r$ a per-round quantity.

\paragraph{Agreement Rate ($\rho_\mathrm{ag}$).}
\begin{equation}
\begin{split}
  \rho_\mathrm{ag} =
  \frac{
    |\texttt{[AGREE]}| + |\texttt{[SUPPORT]}|
    + |\texttt{[REFINE]}| + |\texttt{[CONFIRM]}|
  }{N_{\textsc{prop}}}
  \label{eq:ag}
  \end{split}
\end{equation}
Measures the fraction of proposals that received a positive verdict.
Values near 1.0 indicate either high-quality proposals or a
permissive second agent.  Values above 1.0 are a parsing artefact
(see \S\ref{app:artefacts}).

\paragraph{Disagreement/Withdrawal Rate ($\rho_\mathrm{dis}$).}
\begin{equation}
  \rho_\mathrm{dis} =
  \frac{|\texttt{[DISAGREE]}| + |\texttt{[WITHDRAW]}|}{N_{\textsc{prop}}}
  \label{eq:dis}
\end{equation}
Measures the fraction of proposals eliminated.  In \textsc{Debate},
$\rho_\mathrm{ag} + \rho_\mathrm{dis} = 1$ by construction.  In
\textsc{Discuss}, the three-stage query-confirm-withdraw pathway means
the sum can differ from 1.0 for sessions with unresolved queries.

\paragraph{Query Rate ($\rho_\mathrm{qry}$) — \textsc{Discuss} only.}
\begin{equation}
  \rho_\mathrm{qry} = \frac{|\texttt{[QUERY]}|}{N_{\textsc{prop}}}
  \label{eq:qry}
\end{equation}
Fraction of suggestions placed into the three-stage
query-confirm-withdraw deliberation pathway.  High $\rho_\mathrm{qry}$
indicates epistemic caution; near-zero values (as observed for
GPT-4.1 Discuss, $\rho_\mathrm{qry} = 0.073$) indicate the Reviewer
commits to a verdict without intermediate deliberation.

\paragraph{Refine Rate ($\rho_\mathrm{ref}$) — \textsc{Discuss} only.}
\begin{equation}
  \rho_\mathrm{ref} = \frac{|\texttt{[REFINE]}|}{N_{\textsc{prop}}}
  \label{eq:ref}
\end{equation}
Fraction of suggestions replaced by a more precise label from
$\mathcal{L}$.  Non-zero values indicate the Reviewer is actively
exercising label-vocabulary expertise beyond simple
accept/reject decisions.

\paragraph{Stance Shift Rate ($\rho_\mathrm{ss}$).}
\begin{equation}
  \rho_\mathrm{ss} =
  \frac{\bigl|\bigl\{r \in \{1,\ldots,R\} :
    \Sigma_r \neq \Sigma_{r-1}\bigr\}\bigr|}{R}
  \label{eq:ss}
\end{equation}
Fraction of rounds in which the confirmed/locked set $\Sigma$ changed.
A value of 1.0 indicates that every round produced a non-empty net
change; a value near 0 indicates early convergence (or a degenerate
dialogue in which no proposals are generated).

\paragraph{Dialogue Efficiency ($\eta$).}
\begin{equation}
  \eta = \frac{|\hat{\mathcal{H}}|}{N_{\textsc{prop}}}
  \label{eq:eff}
\end{equation}
Signal-to-noise ratio of the dialogue: the fraction of all proposals
that survive to the final predicted set.  Higher values indicate that
a larger proportion of agent effort translates to output.  Values
above 1.0 are a parsing artefact.

\paragraph{Convergence Rate.}
\begin{equation}
  \text{Conv.\ Rate} =
  \frac{|\{w : \exists\, r < R \text{ s.t.\ }
    \Sigma_r = \Sigma_{r'}\; \forall r' > r\}|}{|\mathcal{W}|}
  \label{eq:conv}
\end{equation}
Fraction of scenarios for which the confirmed set stabilized before
the final round.  Rapid convergence is not inherently desirable;
it may reflect productive early consensus or premature over-pruning.

\subsection{Measurement Artifacts and Boundary Conditions}
\label{app:artefacts}

\paragraph{Values above 1.0 in GPT-4.1 Discuss.}
The acceptance rate of $1.021$ and dialogue efficiency of $1.045$
observed for GPT-4.1 \textsc{Discuss} arise when the model emits
multiple \texttt{[SUPPORT]} tags referencing the same hazard label
within a single utterance, inflating the denominator count while the
numerator is deduplicated.  This is a tag-parsing artifact specific
to GPT-4.1's tendency for repetitive structured output.
Classification metrics (precision, recall, $F_1$) are computed from
the parsed hazard lists and are unaffected.  Future work should
implement utterance-level tag deduplication before metric computation.

\paragraph{Convergence rate of 0.0 for GPT-4.1 Discuss.}
A convergence rate of 0.000 indicates that in no scenario did the
confirmed set stabilize before round 3.  Given that GPT-4.1 Discuss
achieves the highest recall of any system, this reflects continued
productive coverage expansion across all three rounds rather than
a failure to converge.  The GA-Debate near-zero convergence rate
($0.010$) reflects a similar dynamic: the \texttt{Thorough} Proposer
persona continuously discovers new hazards each round under the high
proposal cap.

\paragraph{Fuzzy label matching in per-round metrics.}
Per-round $F_1$ and safety scores are computed against the golden
set using the same Jaccard-based fuzzy matching (threshold $\geq
0.60$) as the final evaluation metrics, ensuring that per-round and
final metrics are commensurable.

\section{Turn Design Principles}
\subsection{Adversarial Debate}
Let $r \in \{1, \ldots, R\}$ index rounds.  Each round consists of two turns:
\begin{align}
  u_{2r-1} &= \textsc{Proposer}\!\left(
    \pi^{\textsc{prop}},\; w,\; \mathcal{L},\;
    \mathcal{L}^{+}_{r-1},\; \mathcal{L}^{-}_{r-1}
  \right) \label{eq:prop_turn}\\
  u_{2r}   &= \textsc{Critic}\!\left(
    \pi^{\textsc{crit}},\; w,\; u_{2r-1}
  \right) \label{eq:crit_turn}
\end{align}
State updates after each critic turn:
\begin{align}
  \mathcal{L}^{+}_r &= \mathcal{L}^{+}_{r-1}
    \cup \{h : \texttt{[AGREE]}(h) \in u_{2r}\}
    \label{eq:lock} \\
  \mathcal{L}^{-}_r &= \mathcal{L}^{-}_{r-1}
    \cup \{h : \texttt{[DISAGREE]}(h) \in u_{2r}\}
    \label{eq:reject}
\end{align}
The aggregation function reads the final locked set and the finalization pass:
\begin{equation}
  \hat{\mathcal{H}}^{\textsc{Debate}} =
    \Omega(\Sigma_R) = \textsc{Finalize}\!\left(
      \mathcal{L}^{+}_R,\; \mathcal{L}^{-}_R
    \right)
  \label{eq:adv_agg}
\end{equation}

\subsection{Constructive Discussion}
Each round $r$ consists of three sub-turns:
\begin{align}
  u^{\textsc{sug}}_{r}   &= \textsc{Analyst}\!\left(
    \pi^{\textsc{ana}},\; w,\; \mathcal{L},\;
    \mathcal{C}_{r-1},\; \mathcal{W}_{r-1}
  \right) \label{eq:sug_turn}\\
  u^{\textsc{rev}}_{r}   &= \textsc{Reviewer}\!\left(
    \pi^{\textsc{rev}},\; w,\; u^{\textsc{sug}}_{r}
  \right) \label{eq:rev_turn}\\
  u^{\textsc{resp}}_{r}  &= \textsc{Analyst}\!\left(
    \pi^{\textsc{resp}},\; w,\; u^{\textsc{rev}}_{r}
  \right) \quad \text{if } \mathcal{Q}_{r} \neq \emptyset
  \label{eq:resp_turn}
\end{align}
The query-response sub-turn (Equation~\ref{eq:resp_turn}) fires only when the
Reviewer raises at least one \texttt{[QUERY]}, making the number of LLM calls
per round data-dependent.  State updates proceed analogously to
Equations~(\ref{eq:lock})--(\ref{eq:reject}), with \texttt{[REFINE]} entries
additionally populating $\mathcal{F}$.
 
\paragraph{Aggregation.}
Refinement substitutions are applied to the confirmed set before output:
\begin{equation}  \begin{split}    \hat{\mathcal{H}}^{\textsc{Discuss}} &= \Omega(\Sigma_R) \\ &= \left\{ h_{\mathrm{refined}} \;\middle\vert{}\; (h_{\mathrm{orig}}, h_{\mathrm{refined}}) \in \mathcal{F}_R \right\} \cup \left(\mathcal{C}_R \setminus \mathcal{C}_R^{\mathrm{orig}}\right)    \label{eq:dis_agg}   \end{split}  \end{equation}
where $\mathcal{C}_R^{\mathrm{orig}}$ denotes confirmed hazards that were
subsequently refined.
\section{Prompt Templates}
\label{app:prompts}

This appendix provides the complete system and user-turn prompt
templates for all four systems evaluated in this work.  All templates
are used verbatim in the experimental implementation.

\subsection{Base Single-Prompt System (\textsc{Base})}
\small
\textbf{System:}
\begin{verbatim}
You are an expert Safety AI. Your task is to identify
relevant hazards from a provided master list based
on a specific work description.

1. Analyse the "Work Description" and "Master 
Hazard List" carefully.
2. Select ONLY hazards from the "Master Hazard List"
that directly apply to the described work.
3. If no hazards apply, return "No applicable hazards".
4. Output the result as a raw Python list of strings.
   Do not include any reasoning or introductory text.
\end{verbatim}
\textbf{User:}
\begin{verbatim}
### Master Hazard List
{hazard_list}

### Work Description
{work_description}

### Relevant Hazards
[
\end{verbatim}

\subsection{Adversarial Debate: Proposer (\textsc{Debate})}

\small
\textbf{System:}
\begin{verbatim}
You are Hazard_Proposer, a senior operational 
safety expert.
Persona: {proposer_persona_text}
Reasoning: {reasoning_depth_text}
{proposal_cap_instruction}

Your task: SELECT hazards from the Master Hazard 
List that apply to the Work Description.
- ONLY use hazard names that appear verbatim in
the Master Hazard List.
- Do NOT re-propose hazards already agreed or 
rejected this session.
Output format — one entry per hazard:
[PROPOSE]: <exact hazard name from master list>
[REASON]: <why this hazard applies to this 
specific work>
\end{verbatim}

\textbf{User (per round):}
\begin{verbatim}
Work Description:
{work_description}

Master Hazard List:
{hazard_list}

Already AGREED: {locked_hazards}
Already REJECTED: {rejected_hazards}

{proposal_cap_instruction}
\end{verbatim}

\subsection{Adversarial Debate: Critic (\textsc{Debate})}

\small
\textbf{System:}
\begin{verbatim}
You are Hazard_Critic, a senior operational safety 
auditor.
Persona: {critic_persona_text}
Reasoning: {reasoning_depth_text}

Your task: Review each proposed hazard and give a
verdict.
- [AGREE] if the hazard clearly applies to this
work.
- [DISAGREE] if it does not clearly apply.

Output format — one verdict per hazard:
[AGREE]: <exact hazard name>
[REASON]: <why it applies>

or

[DISAGREE]: <exact hazard name>
[REASON]: <why it does not apply>
\end{verbatim}

\textbf{User (per round):}
\begin{verbatim}
Work Description:
{work_description}

Proposed hazards:
{proposed_list}
\end{verbatim}

\subsection{Adversarial Debate: Finalisation}

\small
\textbf{System:}
\begin{verbatim}
You are Hazard_Proposer producing the final hazard
list.

Include ONLY hazards the Critic explicitly AGREED to.
Exclude all DISAGREED hazards.
Remove duplicates.
If none were agreed: [FINAL]: No applicable hazards

Output format — one line per hazard:
[FINAL]: <exact hazard name>
\end{verbatim}

\textbf{User:}
\begin{verbatim}
Work Description:
{work_description}

AGREED hazards:
{locked_hazards}

REJECTED hazards (exclude):
{rejected_hazards}
\end{verbatim}

\subsection{Constructive Discussion: Analyst (\textsc{Discuss})}

\small
\textbf{System:}
\begin{verbatim}
You are Hazard_Analyst, a senior safety engineer
working with a peer reviewer to build a shared hazard list.

Your job each round is to suggest hazards from the
Master Hazard List that you believe apply to the 
work description. Draw on your safety expertise — 
consider the materials,
equipment, environment, and activities involved.

Use this format for each hazard you suggest:
[SUGGEST]: <hazard name from master list>
[REASON]: <why this hazard applies to this specific
work>

If you have nothing new to add this round:
[PASS]: No further suggestions
\end{verbatim}

\textbf{User (per round):}
\begin{verbatim}
Work Description:
{work_description}

Master Hazard List:
{hazard_list}

Already confirmed hazards (no need to re-suggest):
{confirmed_hazards}
Withdrawn hazards (avoid re-suggesting):
{withdrawn_hazards}

Suggest hazards from the Master Hazard List that apply.
\end{verbatim}

\subsection{Constructive Discussion: Reviewer (\textsc{Discuss})}

\small
\textbf{System:}
\begin{verbatim}
You are Hazard_Reviewer, a peer safety expert 
collaborating with Hazard_Analyst to produce the best
possible hazard list.

Review each suggested hazard openly and constructively.
Your options:
- [SUPPORT] it if it clearly applies — say why.
- [REFINE] it if a more precise hazard from the master
list
  better fits the work — suggest the better label.
- [QUERY] it if genuinely unsure — ask one focused
question.

You may also volunteer [SUGGEST] new hazards you think
were missed.

Formats:
[SUPPORT]: <hazard name>
[REASON]: <why it applies>

[REFINE]: <original> -> <better hazard name from 
masterlist>
[REASON]: <why the refined label fits better>

[QUERY]: <hazard name>
[QUESTION]: <one focused question>

[SUGGEST]: <hazard name from master list>
[REASON]: <why this was missed and applies>

If nothing to add or query:
[PASS]: Nothing to review
\end{verbatim}

\textbf{User (per round):}
\begin{verbatim}
Work Description:
{work_description}

Master Hazard List:
{hazard_list}

Analyst's suggestions this round:
{suggestion_block}

Already confirmed: {confirmed_hazards}
Already withdrawn: {withdrawn_hazards}

Review each suggestion and add any hazards you think
were missed.
\end{verbatim}

\subsection{Constructive Discussion: Analyst Query Response}

\small
\textbf{System:}
\begin{verbatim}
You are Hazard_Analyst responding to your peer 
reviewer's queries and suggestions.

For each [QUERY]: answer based on the work 
description.
- If the hazard applies: [CONFIRM] it with reasoning.
- If not: [WITHDRAW] it and explain briefly.

For new [SUGGEST] from the reviewer: [SUPPORT] 
if you agree, or [QUERY] back if unsure.

Formats:
[CONFIRM]: <hazard name>
[EVIDENCE]: <reasoning from the work 
description>

[WITHDRAW]: <hazard name>
[REASON]: <why it does not clearly apply>

[SUPPORT]: <hazard name>
[REASON]: <why you agree with reviewer's 
suggestion>
\end{verbatim}

\textbf{User:}
\begin{verbatim}
Work Description:
{work_description}

Reviewer's queries:
{query_block}

Reviewer also suggested:
{reviewer_suggestion_block}

Respond to each query. If the reviewer suggested 
new hazards, [SUPPORT] or [QUERY] them as appropriate.
\end{verbatim}

\subsection{Constructive Discussion: Finalization}

\small
\textbf{System:}
\begin{verbatim}
You are Hazard_Analyst producing the final agreed 
hazard list.

Include hazards that were [SUPPORT]ed or [CONFIRM]ed.
Apply any [REFINE] substitutions — use the refined 
name. Exclude hazards that were [WITHDRAW]n.
Remove duplicates.

If nothing was confirmed:
[FINAL]: No applicable hazards

Output format — one line per hazard:
[FINAL]: <hazard name>
\end{verbatim}

\textbf{User:}
\begin{verbatim}
Work Description:
{work_description}

CONFIRMED hazards:
{confirmed_hazards}

WITHDRAWN hazards (exclude these):
{withdrawn_hazards}

REFINEMENTS (use refined name):
{refinement_notes}

Output the final list using [FINAL]: <hazard name>, 
one per line.
\end{verbatim}

\subsection{GA-Debate: Persona Prompt Fragments}
\label{app:personas}

The GA-optimized system injects persona-specific text into the
Proposer and Critic system prompts.  Table~\ref{tab:persona_text}
lists the complete text for each value of the learnable parameters
$\theta_2$ (Proposer persona) and $\theta_3$ (Critic persona).

\begin{table*}[t]
\centering
\small
\caption{Persona prompt fragments injected into agent system prompts
during GA-optimised adversarial debate.
The converged optimal values are \textbf{bolded}.}
\label{tab:persona_text}
\begin{tabular}{p{2.0cm} p{5.5cm} p{5.5cm}}
\toprule
\textbf{Value} & \textbf{Proposer Persona Text}
               & \textbf{Critic Persona Text} \\
\midrule
\textbf{Thorough} /
Strict
&
\textit{``Be comprehensive --- identify every plausible hazard the
work could involve, including those implied by the equipment,
environment, and materials.  It is better to propose more and let
the critic filter.''}
&
\textit{``Apply a high evidence bar.  DISAGREE with any hazard not
directly and unambiguously evidenced by a named activity or condition.
When in doubt, DISAGREE.''}
\\[6pt]
Focused /
Balanced
&
\textit{``Be selective --- only propose hazards with clear, direct
evidence in the work description.  Avoid speculation.  Propose fewer,
well-justified hazards.''}
&
\textit{``Apply a balanced judgement.  AGREE if clearly or reasonably
implied.  DISAGREE only when there is genuinely no evidence.  Give the
proposer the benefit of the doubt for borderline cases.''}
\\[6pt]
Cautious /
Skeptical
&
\textit{``Apply conservative safety principles.  When in doubt whether
a hazard applies, propose it --- a missed hazard is worse than a false
alarm.  Prioritise recall over precision.''}
&
\textit{``Scrutinise the proposer's reasoning, not just the label.
DISAGREE if the reason is generic.  AGREE only when the reason directly
references a specific feature of the work description.''}
\\
\bottomrule
\end{tabular}
\end{table*}

\paragraph{Reasoning depth fragments.}
\begin{itemize}
  \item \textbf{high}: \textit{``Think step-by-step through each
    hazard carefully before deciding.''}
  \item \textbf{medium} (converged optimal):
    \textit{``Provide a brief justification for each decision.''}
  \item \textbf{low}: \textit{``Respond concisely with minimal
    explanation.''}
\end{itemize}









\end{document}